\documentclass[conference]{IEEEtran}
\IEEEoverridecommandlockouts
\usepackage{cite}
\usepackage{amsmath,amssymb,amsfonts}
\usepackage{algorithmic}
\usepackage{graphicx}
\usepackage{textcomp}
\usepackage{xcolor}
\usepackage{float}
\usepackage{url}
\usepackage{balance}
\usepackage{hyperref}
\usepackage{placeins}

\def\BibTeX{{\rm B\kern-.05em{\sc i\kern-.025em b}\kern-.08em
    T\kern-.1667em\lower.7ex\hbox{E}\kern-.125emX}}
\begin{document}

\title{
    GPS-DRIFT: Marine Surface Robot Localization using IMU-GPS Fusion and Invariant Filtering\\[0.5em]}

\author{
\IEEEauthorblockN{Surya Pratap Singh, Tsimafei Lazouski, and Maani Ghaffari}
\thanks{S. P. Singh, T. Lazouski, and M. Ghaffari are with the Department of Naval Architecture and Marine Engineering at the University of Michigan, Ann Arbor, MI 48109, USA. \texttt{\{suryasin,tlazousk,maanigj\}@umich.edu}.}
}

\maketitle

\begin{abstract}
This paper presents an extension of the DRIFT invariant state estimation framework, enabling robust fusion of GPS and IMU data for accurate pose and heading estimation. Originally developed for testing and usage on a marine autonomous surface vehicle (ASV), this approach can also be utilized on other mobile systems. Building upon the original proprioceptive only DRIFT algorithm, we develop a symmetry-preserving sensor fusion pipeline utilizing the invariant extended Kalman filter (InEKF) to integrate global position updates from GPS directly into the correction step. Crucially, we introduce a novel heading correction mechanism that leverages GPS course-over-ground information in conjunction with IMU orientation, overcoming the inherent unobservability of yaw in dead-reckoning. The system was deployed and validated on a customized Blue Robotics BlueBoat, but the methodological focus is on the algorithmic approach to fusing exteroceptive and proprioceptive sensors for drift-free localization and reliable orientation estimation. This work provides an open source solution for accurate yaw observation and localization in challenging or GPS-degraded conditions, and lays the groundwork for future experimental and comparative studies.
\end{abstract}

\section{Introduction}
Accurate state estimation is fundamental to the reliable autonomous navigation of marine surface vehicles. Robust localization typically requires the combination of high-rate inertial measurements from IMUs with global position updates from the Global Positioning System (GPS)~\cite{fossen2021, groves2013}. Although GPS provides global position information, its low update rate, vulnerability to outages, and measurement noise require dead-reckoning via IMU integration between GPS fixes. However, inertial-only dead-reckoning results in unbounded drift in both position and orientation. In particular, yaw (heading) is fundamentally unobservable in pure inertial navigation due to the lack of an external reference for rotations about the gravity axis~\cite{barrau2016}.

The DRIFT (Dead Reckoning in Field Time) framework addresses these challenges through a symmetry-preserving state estimation approach based on the invariant extended Kalman filter (InEKF)~\cite{lin2024}. The original DRIFT algorithm uses the Lie group theory to achieve robust and consistent odometry for legged and wheeled robots, but direct fusion of GPS for position and heading correction, particularly in marine environments, remains an active area of research.

In this work, we extend the DRIFT framework to support GPS/IMU fusion for marine surface vehicles, with an emphasis on heading (yaw) correction. We present an algorithmic pipeline that integrates GPS updates into the InEKF correction step and introduces a novel mechanism to leverage GPS position fixes to resolve yaw ambiguity.

\begin{figure}[t]
    \centering
    \includegraphics[width=0.7\linewidth]{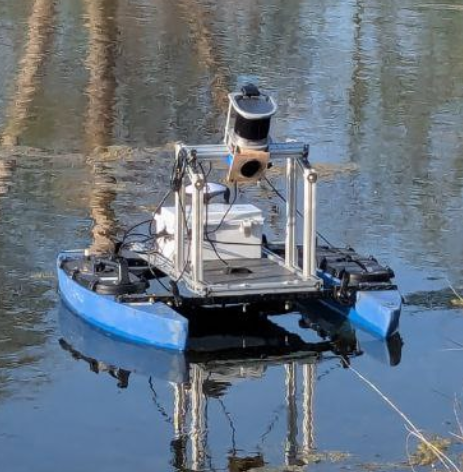}
    \includegraphics[width=0.7\linewidth]{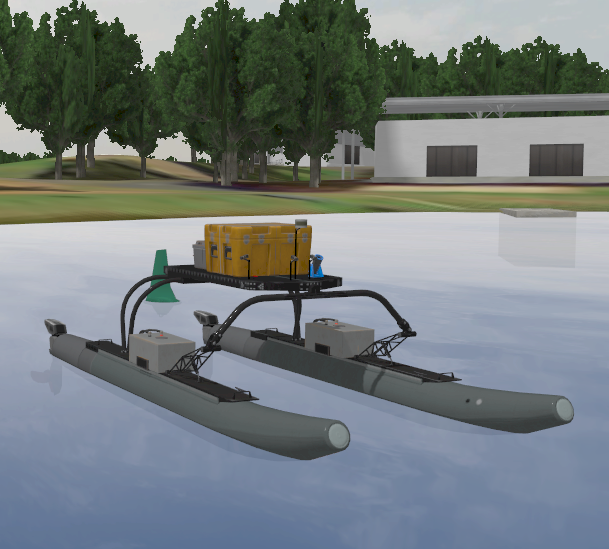}
    \caption{The top image demonstrates the customized BlueBoat equipped with an real time kinematic (RTK), centimeter level accuracy, dual GPS and Vectonav VN-100 IMU. The bottom image is a screenshot of the WAMV vessel from the VRX Gazebo marine environment~\cite{vrx}.}
    \label{fig:PlaceHolder}
\end{figure}

 Our open source implementation has been validated in a simulated marine environment physics-accurate, Gazebo, Virtual RobotX (VRX), on the default WAMV vessel equipped with both an IMU and GPS~\cite{vrx}. Real-world validation is conducted on a customized BlueRobotics BlueBoat platform, equipped with a real time kinematic (RTK), centimeter level accuracy, dual GPS setup, along with a Vectornav VN-100 IMU. In general, this work has shown substantial improvements in localization and accurate orientation calculation.

The main contributions of this work are as follows.
\begin{enumerate}
\item Extension of the DRIFT framework to support GPS/IMU sensor fusion for marine surface vehicles
\item Development of a novel heading correction mechanism utilizing GPS course-over-ground information within the invariant filtering framework
\item Implementation and validation of the algorithm in simulated and field environments.
\item Open-source ROS2 implementation that enables reproducible research and practical deployment: \href{https://github.com/UMich-CURLY/ASV_localization}{GPS-DRIFT Github}
\end{enumerate}

\section{Related Work}

\subsection{Marine State Estimation}
Robust state estimation is fundamental for autonomous surface vessels. Traditionally, localization has used Extended Kalman Filter (EKF) sensor fusion to combine high-rate inertial measurements from IMUs with low-rate globally referenced updates from exteroceptive sensors~\cite{gpimuekf}.

Although commonly used in robotics, standard EKF-based methods suffer from significant limitations when applied to marine environments. These approaches are susceptible to performance degradation in highly non-linear regimes or poorly observable conditions due to their reliance on local linearization of the current state estimate. Linearization errors can accumulate over time, leading to filter divergence and reduced estimation accuracy~\cite{ekfrules}. 

\subsection{Invariant Extended Kalman Filtering}

To address these limitations, recent advances have introduced symmetry-preserving observers based on Lie group theory. The invariant extended Kalman filter (InEKF) models the robot pose and velocity in matrix Lie groups, allowing the linearization of the filter to be independent of the estimated state~\cite{barrau2016}. This approach leads to improved convergence and consistency, particularly for systems with non-linear error dynamics. The DRIFT framework~\cite{lin2024} is a notable implementation of InEKF for real-time robot state estimation, originally developed for legged and wheeled robots, with modular support for multiple sensor modalities.

\subsection{Marine Surface Vehicle Localization Challenges}
In marine robotics, state estimation poses unique challenges. Surface and underwater vehicles often operate in GPS-denied environments or where GNSS signals are intermittent or degraded~\cite{marInEKF}. While GPS provides global position fixes, its low update rate and susceptibility to noise and outages make it insufficient for reliable state tracking alone. Dead-reckoning with IMU data can provide high-frequency pose updates, but integration errors accumulate rapidly, especially in heading (yaw), which is unobservable from accelerometer data alone due to the indistinguishability of rotation about the gravity vector~\cite{martinelli}.

\subsection{Existing Solutions and Research Gaps}
Alternative methods—including DVL-based velocity estimation, vision/acoustic SLAM, and magnetometer fusion—have limitations such as high cost, environmental sensitivity, and vulnerability to magnetic disturbances~\cite{MarVesOver}. Multi-sensor fusion approaches can help, but often require complex and expensive sensor suites. However, reliable heading correction in marine surface vehicles via GPS/IMU fusion remains underexplored, particularly using the invariant filtering framework. This paper builds upon the DRIFT methodology to develop an open-source, ROS2-compatible solution for integrating GPS-based heading corrections with IMU-driven dead-reckoning, targeting robust and drift-free localization for marine robots.

\section{Background} 
The DRIFT framework utilizes an Invariant EKF (InEKF), which models robot states and errors using matrix Lie groups and their associated Lie algebras.
For detailed usage of InEKF for localization and its specific implementation within the DRIFT framework on which our work is based, we refer the reader to the original paper ~\cite{lin2024}. Yet, a summary of the core concepts needs to be presented to effectively communicate the contributions of our work.

\subsection{DRIFT and InEKF}
The Invariant Extended Kalman Filter (InEKF) used in DRIFT models the ASV’s pose, velocity, and biases on the matrix Lie group $\mathrm{SE}_2(3)$, with state
\begin{equation}
  X \;=\;\bigl(R,\;v,\;p,\;b_\omega,\;b_a\bigr) \in \mathrm{SE}_2(3) \times \mathbb{R}^6,
\end{equation}
where:
\begin{align*}
R &\in \mathrm{SO}(3): \text{ 3D orientation matrix} \\
v, p &\in \mathbb{R}^3: \text{ body-frame velocity and inertial-frame position} \\
(b_\omega, b_a) &\in \mathbb{R}^6: \text{gyroscope and accelerometer biases}
\end{align*}

The InEKF framework represents uncertainty in the tangent space (Lie algebra) of each manifold, enabling consistent linearization that is independent of the current state estimate. This property is crucial to maintain the consistency of the filter in marine environments where large orientation changes are common.~\cite{barrau2016,lin2024}.

The standard InEKF operation alternates between two phases:

\textbf{a. Prediction Step:} State propagation using IMU measurements (angular velocity $\omega_m$ and linear acceleration $a_m$) over time interval $\Delta t$:
\begin{align}
\frac{d}{dt} \bar{X}_t &= f_u(\bar{X}_t) \\
\frac{d}{dt} P^l_t &= A^l_t P^l_t + P^l_t {A^l_t}^\mathsf{T} + Q_t
\end{align}
where $\bar{X}_t \in \mathrm{SE}_2(3)$ represents the left-invariant state estimate, $P^l_t$ denotes the covariance matrix, $f_u(\cdot)$ describes the IMU-driven dynamics, $A^l_t$ represents the left-invariant Jacobian, and $Q_t$ represents process noise.

\textbf{b. Correction Step:} At discrete measurement times $t_k$, the state update follows:
\begin{align}
\bar{X}_{t_k}^+ &= \bar{X}_{t_k} \exp(L^l_{t_k}(\bar{X}_{t_k}^{-1}Y_{t_k} - b)^\wedge) \\
P^{l+}_{t_k} &= (I - L^l_{t_k}H)P^l_{t_k}(I - L^l_{t_k}H)^\mathsf{T} + L^l_{t_k}\bar{N}_k {L^{l}}^\mathsf{T}_{t_k}
\end{align}
where $H$ represents the measurement Jacobian, $\bar{N}_k$ denotes measurement noise, $Y_{t_k}$ represents the raw measurement on the group, and $b$ represents a constant bias vector.

Full details and derivation of this method can be traced in~\cite{lin2024}.

\section{GPS Fusion and Heading Correction within DRIFT} 

The proposed extension integrates GPS position measurements into the left-invariant framework while simultaneously enabling heading correction. The GPS configuration on the BlueBoat platform provides centimeter-level accuracy position readings, which are incorporated through the measurement model.

\subsection{Position Measurement Integration}

Since the GPS position measurements are in the world frame, we follow the left-invariant observation model ~\cite[Eq.~(7)]{lin2024}:
\begin{equation}
{Y}_t = {X}_t {b} + {V}_t
\end{equation}

The incoming position measurements can be written as:
\begin{equation}
\widetilde{{P}_{t}} = {P}_t + \omega_t^P
\end{equation}
where $\widetilde{{P}_{t}}$ is the measured position, ${P_t}$ is the true position and $\omega_t^P$ is measurement noise. With these formulations, we can expand the equation to take the form:
\begin{equation}
\underbrace{\begin{bmatrix}
\widetilde{P}_t \\[2pt]
0 \\[2pt]
1
\end{bmatrix}}_{Y_t}
=
\underbrace{\begin{bmatrix}
R_t & v_t & P_t \\[6pt]
0   & 1   & 0   \\[2pt]
0   & 0   & 1
\end{bmatrix}}_{X_t}
\underbrace{\begin{bmatrix}
0 \\[2pt]
0 \\[2pt]
1
\end{bmatrix}}_{b}
\;+\;
\underbrace{\begin{bmatrix}
w_t^P \\[2pt]
0     \\[2pt]
0
\end{bmatrix}}_{V_t}
\end{equation}

Now, as per the left-invariant observation model, we need to find the innovation term ${Z}$:
\begin{equation}
Z = X^{-1}Y - b
\end{equation}
\begin{equation}
Z_t = 
\begin{bmatrix}
  R_t^\mathsf{T} & -R_t^\mathsf{T} v_t & -R_t^\mathsf{T} P_t \\
  0        & 1             & 0            \\
  0        & 0             & 1
\end{bmatrix}
\begin{bmatrix}
  \widetilde{P}_t \\[4pt]
  0 \\[2pt]
  1
\end{bmatrix}
-
\begin{bmatrix}
  0 \\[2pt]
  0 \\[2pt]
  1
\end{bmatrix}
\end{equation}

which can be expressed as:

\begin{equation}
Z_t = R_t^\mathsf{T} \bigl(\widetilde{P}_t - P_t \bigr)
\end{equation}

Next, we formulate the measurement Jacobian ${H}$. From ~\cite[Eq.~(9, 10)]{lin2024} we know that:
\begin{equation}
H\,\zeta = \zeta^{\wedge}\,b
\end{equation}

Which can be expressed in the following way:
\begin{equation}
H_t
\begin{bmatrix}
  \zeta_t^{\omega} \\[2pt]
  \zeta_t^{v}      \\[2pt]
  \zeta_t^{P}
\end{bmatrix}
=
\begin{bmatrix}
  \zeta_t^{P} \\[2pt]
  0           \\[2pt]
  0
\end{bmatrix}
\end{equation}

Ultimately leading to the following Jacobian measurement matrix $3\times15$:
\begin{equation}
H_t =
\begin{bmatrix}
  0_{3\times3} & 0_{3\times3} & I_{3} & 0_{3\times3} & 0_{3\times3}
\end{bmatrix}
\end{equation}

Finally, the measurement covariance ${N}$, in~\cite[Eq.~(1)]{lin2024} can be expressed as:

\begin{equation}
{N}_t = X_t^{-1} \, \mathrm{Cov}(V_t) \, X_t^{-\top}
\end{equation}

For GPS position noise \( w_t^P \in \mathbb{R}^{3 \times 3} \), with

\begin{equation}
X_t^{-1} =
\begin{bmatrix}
R_t^{-1} & -R_t^{-1}v_t & -R_t^{-1}P_t \\
0 & 1 & 0 \\
0 & 0 & 1
\end{bmatrix},
\end{equation}

\begin{equation}
\quad
\mathrm{Cov}(V_t) =
\begin{bmatrix}
w_t^P \\ 0 \\ 0
\end{bmatrix},
\end{equation}

Neglecting the translation terms, we arrive at the following:
\begin{equation}
{N}_t = R_t^\mathsf{T} w_t^P R_t
\end{equation}
using the identity \( R_t^{-1} = R_t^\mathsf{T} \).

\subsection{Orientation Correction Integration}

The IMU provides direct orientation measurements in the form of a rotation matrix $\widetilde{R}_t$. The measurements are inherently subject to drift; however, they can be corrected using positional information from GPS. Using the known relationship between consecutive GPS positions, heading references are generated for correction.

The incoming orientation measurements for each basis vector $(e_1, e_2, e_3)$ can be written as:
\begin{equation}
\widetilde{R}_{ti} = R_t e_i + w_t^R, \quad i = 1, 2, 3
\end{equation}
Following the left-invariant observation model:
\begin{equation}
\underbrace{\begin{bmatrix}
\widetilde{R}_{ti} \\[2pt]
0 \\[2pt]
0
\end{bmatrix}}_{Y_{ti}}
=
\underbrace{\begin{bmatrix}
R_t & v_t & P_t \\[6pt]
0   & 1   & 0   \\[2pt]
0   & 0   & 1
\end{bmatrix}}_{X_t}
\underbrace{\begin{bmatrix}
e_i \\[2pt]
0 \\[2pt]
0
\end{bmatrix}}_{b_i}
\;+\;
\underbrace{\begin{bmatrix}
w_t^R \\[2pt]
0     \\[2pt]
0
\end{bmatrix}}_{V_{ti}}
\end{equation}

The innovation terms for each basis vector are computed as:
\begin{equation}
Z_{ti} = X_t^{-1}Y_{ti} - b_i = R_t^\mathsf{T} \widetilde{R}_{ti} - e_i, \quad i = 1, 2, 3
\end{equation}
GPS measurements refine the overall pose estimation, and the corrected pose subsequently improves orientation estimates through the InEKF structure.

The measurement Jacobian for each orientation component is derived from $H_i \zeta = \zeta^{\wedge} b_i$:

\begin{equation}
H_{ti}
\begin{bmatrix}
  \zeta_t^{\omega} \\[2pt]
  \zeta_t^{v}      \\[2pt]
  \zeta_t^{P}
\end{bmatrix}
=
\begin{bmatrix}
  -\text{skew}(e_1)\cdot \zeta_t^{\omega} \\[2pt]
  0           \\[2pt]
  0
\end{bmatrix}
\end{equation}

which results in:

\begin{equation}
H_{ti} = [-\text{skew}(e_i) \quad 0_{3 \times 12}], \quad i = 1, 2, 3
\end{equation}
where $\text{skew}(\cdot)$ represents the skew-symmetric matrix operator.

The measurement covariance for orientation follows:
\begin{equation}
N_{ti} = X_t^{-1}\,\mathrm{Cov}(V_{ti})\,X_t^{-\mathsf{T}} = R_t^\mathsf{T}\,w_t^R\,R_t
\end{equation}

\subsection{Combined Pose Correction}

GPS positional updates and IMU orientation measurements are stacked together in a unified pose correction step. This ensures simultaneous corrections to both position and orientation. The stacking innovation vector is the following.
\begin{equation}
Z_t = \begin{bmatrix} R_t^\mathsf{T}(\tilde{p}_t - p_t) \\ R_t^\mathsf{T} \tilde{R}_t e_1 - e_1 \\ R_t^\mathsf{T} \tilde{R}_t e_2 - e_2 \\ R_t^\mathsf{T} \tilde{R}_t e_3 - e_3 \end{bmatrix} \in \mathbb{R}^{12 \times 1}
\end{equation}

The combined measurement Jacobian becomes:
\begin{equation}
H = \begin{bmatrix}
0_{3 \times 3} & 0_{3 \times 3} & I_3 & 0_{3 \times 3} & 0_{3 \times 3} \\
-\text{skew}(e_1) & 0_{3 \times 3} & 0_{3 \times 3} & 0_{3 \times 3} & 0_{3 \times 3} \\
-\text{skew}(e_2) & 0_{3 \times 3} & 0_{3 \times 3} & 0_{3 \times 3} & 0_{3 \times 3} \\
-\text{skew}(e_3) & 0_{3 \times 3} & 0_{3 \times 3} & 0_{3 \times 3} & 0_{3 \times 3}
\end{bmatrix} \in \mathbb{R}^{12 \times 15}
\end{equation}

The combined measurement covariance matrix is:
\begin{equation}
\begin{split}
N = \mathrm{diag}\big[&R_t^\mathsf{T} \mathrm{Cov}(\text{pos}) R_t, \\
                      &R_t^\mathsf{T} \mathrm{Cov}(\text{rot}) R_t, \\
                      &R_t^\mathsf{T} \mathrm{Cov}(\text{rot}) R_t, \\
                      &R_t^\mathsf{T} \mathrm{Cov}(\text{rot}) R_t \big]
\end{split}
\end{equation}

\subsection{Heading Correction Mechanism}

The key innovation of this work lies in the heading correction mechanism that leverages GPS position fixes. Traditional dead-reckoning approaches suffer from yaw unobservability, but the proposed method exploits the relationship between heading correction achieved through the use of GPS position inputs and IMU orientation estimates through the left-invariant framework.

Heading correction is achieved through the orientation component of the combined pose correction framework. By incorporating the corrected orientation estimates into the rotation measurement model following the formulation $Y_t = X_t b + V_t$, the system can effectively resolve the yaw ambiguity while maintaining consistency with the invariant filtering framework.

The InEKF correction update is performed using the standard left-invariant correction:
\begin{equation}
\bar{X}_{t_k}^+ = \bar{X}_{t_k} \exp(L^l_{t_k} Z_t^{\wedge})
\end{equation}
where $L^l_{t_k} = (H^\mathsf{T} N^{-1} H)^{-1} H^\mathsf{T} N^{-1}$ is the left-invariant gain matrix.

This formulation ensures that position corrections automatically contribute to heading corrections through the coupling inherent in the $\mathrm{SE}_2(3)$ group structure, while explicit orientation measurements provide additional constraints for yaw estimation.

\section{Experimental Setup and Results}

To validate the proposed GPS/IMU fusion extension to the DRIFT framework, we conducted experiments both in a high-fidelity simulation environment and on a physical marine surface vehicle.

\subsection{ASV Platform Overview}
The experimental platform is a customized Blue Robotics BlueBoat, a compact autonomous surface vessel (ASV) equipped with a precise GPS, a 9-axis IMU, and an onboard computer running ROS2. This configuration enables high-frequency inertial measurements and centimeter-level GPS updates, making it well suited for testing the proposed invariant filtering approach. While the primary results presented here focus on the simulation environment, preliminary deployment on the physical BlueBoat platform has been completed, with full validation ongoing.

\subsection{Simulation Experiments}
We used the Gazebo-based Virtual RobotX (VRX) simulation environment to evaluate the performance of the localization pipeline. This platform simulates realistic marine dynamics and provides:

\begin{itemize}
    \item Ground-truth odometry for benchmarking
    \item Realistic GPS and IMU sensor models
    \item A configurable marine environment for varied scenarios
\end{itemize}

The DRIFT InEKF node was configured to subscribe to both the GPS and IMU topics. The results of the localization were then cross-evaluated against the ground truth (GT) measurements and simple GPS-tracking. The heading from the modified DRIFT algorithm was compared with the data directly from the ground truth odometry topic.

\subsection{Results}

\begin{figure}[t]
    \centering
    \includegraphics[width=0.8\linewidth]{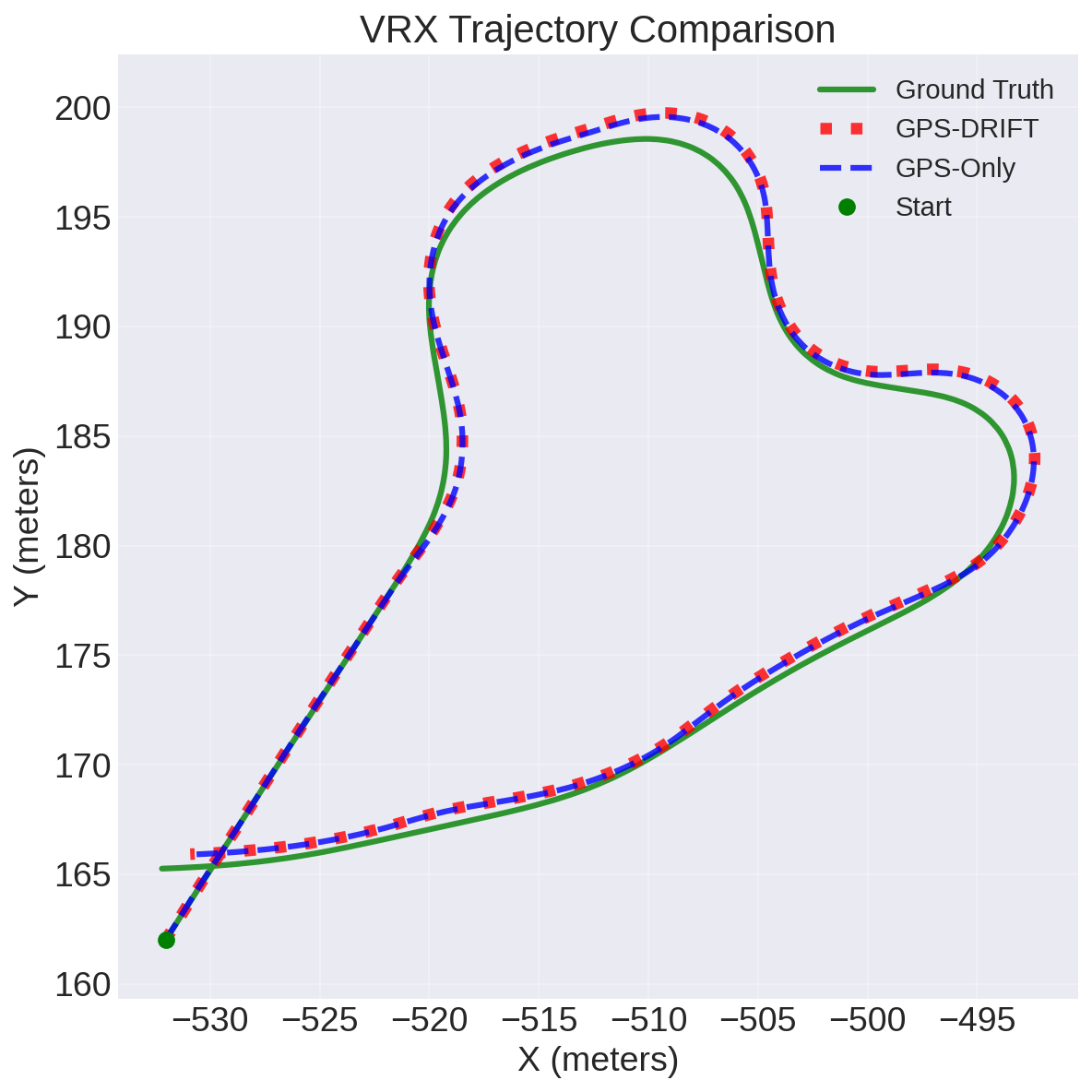}
    \caption{Trajectory of the WAMV ASV in the VRX Gazebo environment.}
    \label{fig:trajectories}
\end{figure}

Figure~\ref{fig:trajectories} shows the route the simulated WAMV boat traveled in the Gazebo environment. We can see that while the localization from the DRIFT algorithm aligned extremely well with the path tracked from the GPS data, both trajectories were offset in comparison to the ground truth. This is largely due to the artificial noise present in both the IMU and GPS readings within the Gazebo simulation. Consequently, such a positional misalignment of the GPS and noise from the IMU will impact the accuracy of the heading correction.

\begin{figure}[t]
    \centering
    \includegraphics[width=0.9\linewidth]{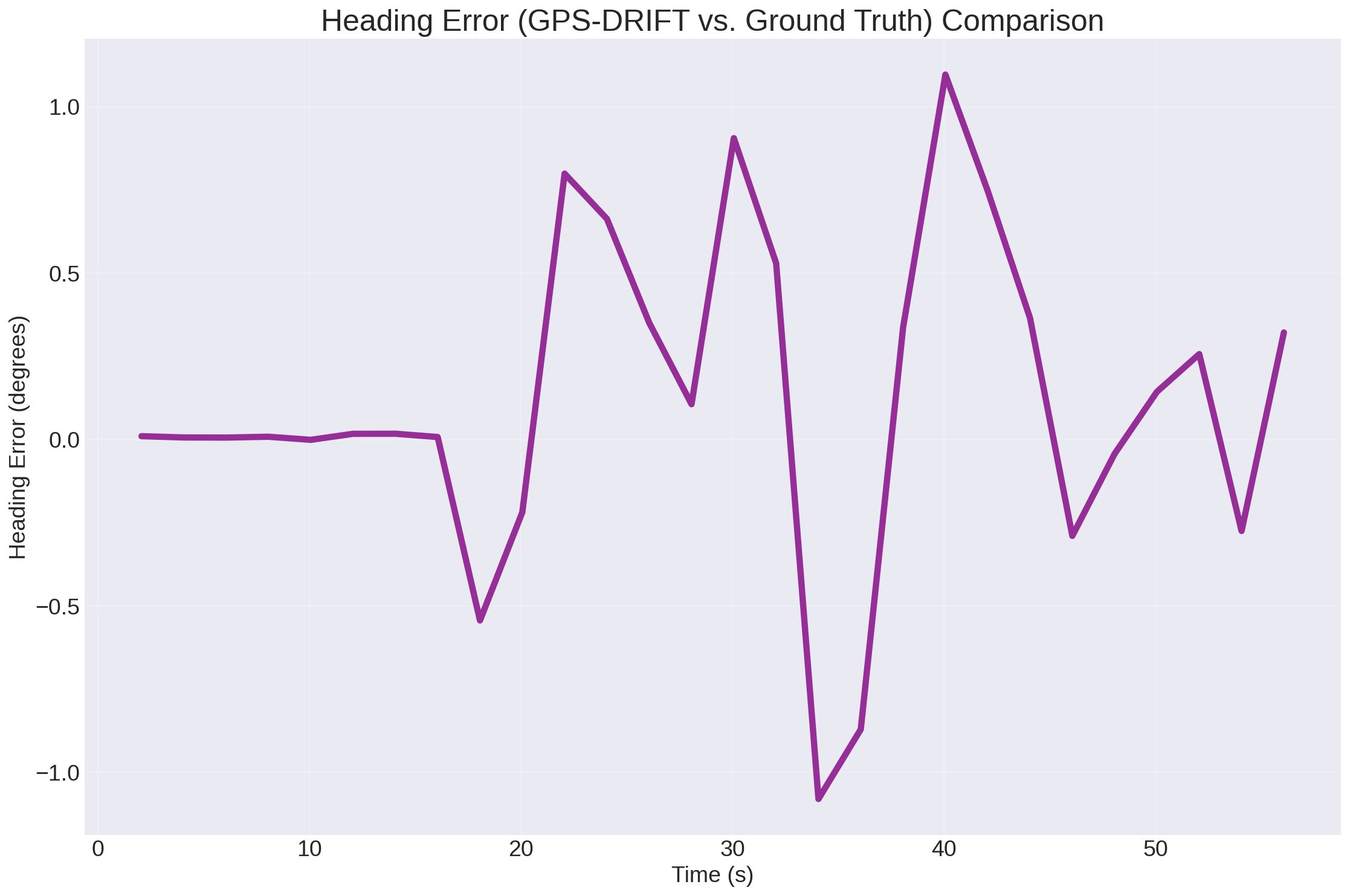}
    \caption{Heading error between output of the DRIFT heading correction output and the ground truth odometry reading}
    \label{fig:head_err}
\end{figure}

Figure~\ref{fig:head_err} shows the heading error between the heading coming from the estimated pose DRIFT calculation and the ground-truth heading readings. From the plot, we can see that following the trajectory shown in Figure~\ref{fig:trajectories}, the maximum heading error consistently remained within within 1 degree. Additionally, with a more accurate alignment between GPS and ground truth, and a less noisy IMU, the heading error would be further minimized, which we aim to observe in the field test results.

\begin{table}[t]
    \centering
    \caption{Comparison of Localization Errors versus Ground Truth for GPS-DRIFT and GPS-Only}
    \begin{tabular}{lcc}
        \hline
        \textbf{Error Metric (vs. GT)} & \textbf{GPS-DRIFT} & \textbf{GPS-Only} \\
        \hline
        Position RMSE (m)         & 1.131 & 1.012 \\
        Position Max Error (m)    & 1.77 & 1.593 \\
        Heading RMSE (deg)        & 0.498  & ---   \\
        Heading Max Error (deg)   & 1.097 & ---   \\
        \hline
    \end{tabular}
    \label{tab:result_summary}
\end{table}

%\begin{table}[t]
%    \centering
%    \caption{Localization Error for GPS-DRIFT vs GPS-Only}
%    \begin{tabular}{lcc}
%        \hline
%        Position RMSE (m)         & 0.059 
%        \\
%        Position Max Error (m)    & 0.329 
%        \\
%    \end{tabular}
%    \label{tab:gps_vs_drift}
%\end{table}

Quantitative results are summarized in Table~\ref{tab:result_summary}, reporting root mean square errors (RMSE) in both position and heading for GPS-DRIFT and GPS-Only compared to ground truth in multiple trials. Note that the performance of GPS-DRIFT localization is heavily impacted by the accuracy of the VRX GPS itself. The RMSE between the GPS-DRIFT and GPS-Only trajectories is 0.059 meters, indicating the overall efficacy of the localization method. 

These results confirm that incorporating GPS position inputs for heading correction allows decent orientation estimation and suppresses long-term drift, even when the GPS position fixes used do not exactly align with the ground truth position seen in the simulation.

\subsection{Preliminary Physical Deployment}

The initial deployment of the system on the BlueBoat ASV has been completed. Sensor integration and real-time localization with GPS / IMU fusion are operational. Data collection for quantitative evaluation is ongoing and will be included in future work. 

\begin{figure}[t]
    \centering
    \includegraphics[width=0.75\linewidth]{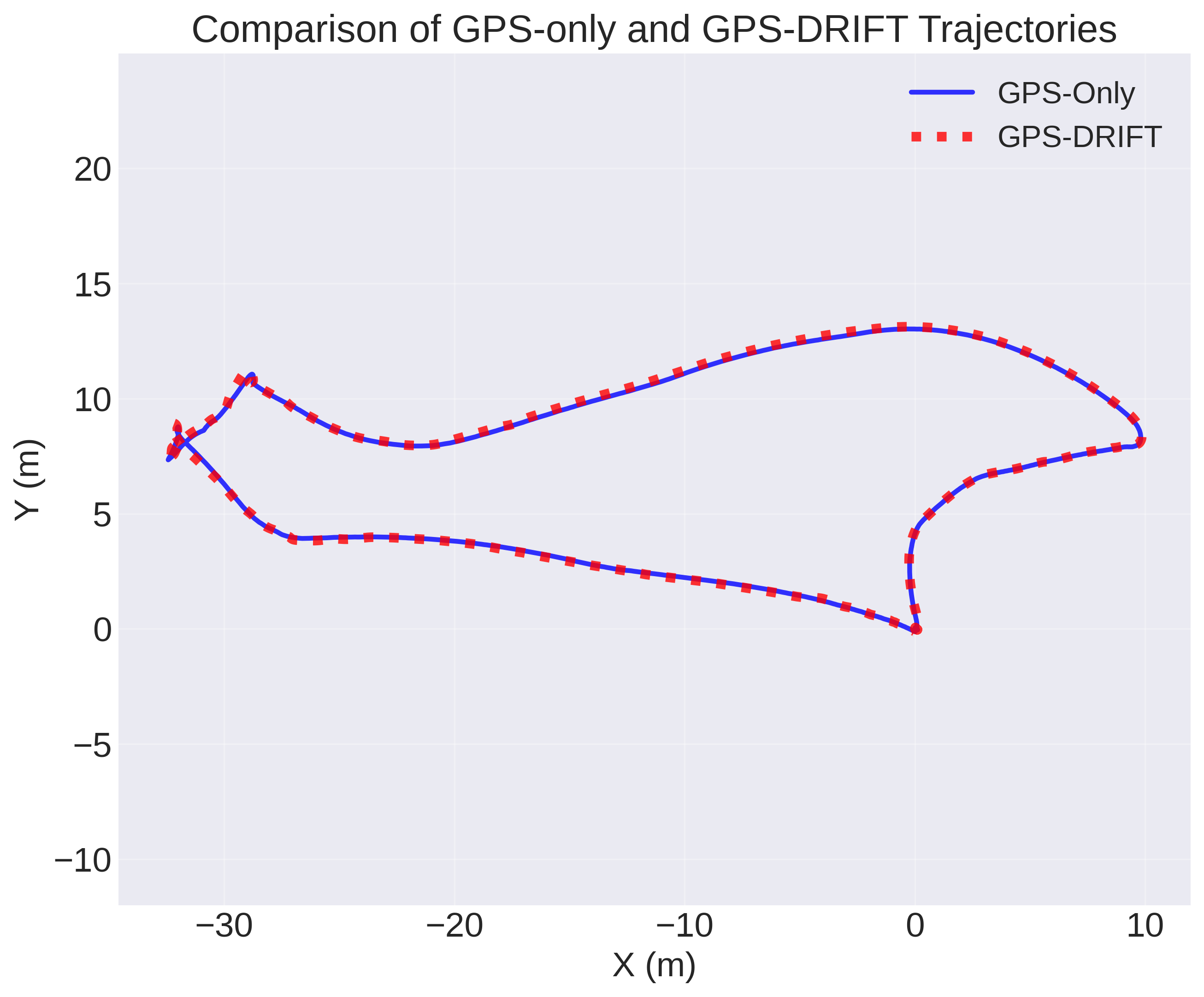}
    \caption{The figure demonstrates an arbitrary trajectory at a pond test with an overlay of the GPS-only trajectory and DRIFT estimated trajectory.}
    \label{fig:ft_res}
\end{figure}

The preliminary pond test results can be seen in Figure~\ref{fig:ft_res}. The preliminary results are extremely promising, as the maximum absolute position error in the DRIFT position output vs. the GPS position is \textless 0.1 meters. As more pond tests are conducted are conducted, the heading correction results will be more accurately validated.

\section{Conclusions and Future Work}
This paper introduced an extension of the DRIFT invariant state estimation framework, focusing on integrating GPS and IMU data to achieve orientation correction for marine surface vehicles. Our proposed approach leverages the InEKF to fuse GPS position readings with IMU data, addressing the inherent yaw unobservability problem in inertial navigation systems. Experimental validation, performed both in the VRX Gazebo simulation and preliminary real-world deployment on the lab's modified Blue Robotics BlueBoat, demonstrated improved localization accuracy and reliable heading estimation. The heading error between the localization estimate and ground truth readings consistently remained within 1 degree in the tests within the VRX environment. The open-source ROS2 implementation supports reproducible research and facilitates practical adoption across various autonomous mobile platforms.

Future efforts will focus on further real-world testing and validation of the system's performance. We also aim to investigate performance when using a cost-effective, standalone GPS unit, and we seek to run longer tests to understand the robustness of the heading correction methodology.

\FloatBarrier 
{\balance
\bibliographystyle{ieeetr}
\bibliography{bibliography.bib}
}

\end{document}